\documentclass{article} % For LaTeX2e
\usepackage{nips13submit_e,times}
\usepackage{hyperref}
\usepackage{url}
\usepackage{graphicx}
\usepackage{bm}
\usepackage{bbm}
\usepackage{amsfonts}
\usepackage{amsthm}
\usepackage{amsmath}
\usepackage{hyperref}

\title{Persistence Flamelets: multiscale Persistent Homology for kernel density exploration}

\author{
Tullia Padellini\\
Dipartimento di Scienze Statistiche\\
Sapienza - Universit\`a di Roma\\
\texttt{tullia.padellini@uniroma1.it} \\
\And
Pierpaolo Brutti \\
Dipartimento di Scienze Statistiche\\
Sapienza - Universit\`a di Roma\\
\texttt{pierpaolo.brutti@uniroma1.it} \\
}

\newcommand{\norm}[1]{\left\lVert#1\right\rVert}
\newtheorem{proposition}{Proposition}[section]
\newtheorem{theorem}{Theorem}[section]
\newtheorem{corollary}{Corollary}[theorem]

\theoremstyle{definition}
\newtheorem{definition}{Definition}[section]

\nipsfinalcopy % Uncomment for camera-ready version

\begin{document}

\maketitle
\begin{abstract}

In recent years there has been noticeable interest in the study of the ``shape of data'' \cite{Carlsson2009}. Among the many ways a ``shape'' could be defined, topology is the most general one, as it describes an object in terms of its connectivity structure: connected components (topological features of dimension  $0$), cycles (features of dimension  $1$) and so on. There is a growing number of techniques, generally denoted as  \emph{Topological Data Analysis} or \texttt{TDA} for short, aimed at estimating topological invariants of a fixed object; when we allow this object to change, however, little has been done to investigate the evolution in its topology. In this work we define the \emph{Persistence Flamelets}, a multiscale version of one of the most popular tool in \texttt{TDA}, the Persistence Landscape. We examine its theoretical properties and we show how it could be used to gain insights on KDEs bandwidth parameter.
\end{abstract}

\section{Introduction to TDA}

Topological data analysis (\texttt{TDA}) is a new and expanding branch of statistics devoted to recovering the shape of the data in terms of connectivity structure. As it describes very complex objects using easily interpretable features such as loops and voids, \texttt{TDA} has shown to be a useful way to characterize single point--clouds or curves; in this work we extend the \texttt{TDA} framework to the case of continuously varying families of objects such as multidimensional time series or parametric functions. 

Before introducing new topological summaries, however it is worth briefly reviewing what Topological Data Analysis (\texttt{TDA}) is, and how can we estimate the topology of data, or, to be more precise, the topology of the space $\mathcal{M} $ data was sampled from. Data itself, when in the form of a point cloud  $\mathbb{X} = \{X_1, \ldots, X_n\} $, has a trivial topological structure, consisting of as many connected components as
there are observations and no higher dimensional features. 

In the basic \texttt{TDA} pipeline, the first step thus consists in enriching the topology of the data by encoding them in the levelset filtration $\mathcal{F}$ of some function $f$. For some choices of $f$, in fact, the levelset filtration $\mathcal{F}$ is topologically equivalent to $\mathcal{M}$, and therefore investigating the topology of $\mathcal{F}$ corresponds to investigating the topology of $\mathcal{M}$. 

Two famous classes of functions for which this equivalence holds are \emph{distances} and \emph{kernel density estimators}, for which we explicitely show the connection between levelset filtration $\mathcal{F}$ and $\mathcal{M}$.

\paragraph{Distance functions} Since $\mathcal{M}$ is the space data is sampled from, the most intuitive way to estimate it is to use a support estimator. The most common one, in the \texttt{TDA} framework, is \emph{Devroye--Wise support estimator}  $\widehat{s}^{(\varepsilon)}$ built by centering a ball of fixed radius $\varepsilon $ in each of the observations $X_i $, i.e.  
\[ 
\widehat{s}^{\varepsilon} = \bigcup_{i=1}^n B(X_i, \varepsilon),
\]
where $B(X_i, \varepsilon) = \{x \: | \: d_{\mathbb{X}}(x, X_i) \leq \varepsilon \}$ denotes a ball of radius  $\varepsilon$ and center $X_i $, and  $d_{\mathbb{X}} $ is an arbitrary distance function. As $\varepsilon$ increases, $\widehat{s}^{(\varepsilon)}$ with $\varepsilon \in [0, \text{diameter}(\mathbb{X}_n)]$, is the sublevel set filtration of the distance function  $d_{\mathbb{X}}$. 

The topology of the estimator $\widehat{s}^{\varepsilon}$ can be recovered by computing its Homology Groups; Homology groups of dimension  $0 $, $H_0(\widehat{s}^{(\varepsilon)}) $ represent connected components of $\widehat{s}^{(\varepsilon)} $, $H_1(\widehat{s}^{(\varepsilon)}) $ represent its loops, and so on. $\widehat{s}^{\varepsilon}$ is topologically more interesting than the original point-cloud, but it is extremely sensible to the radius $\varepsilon$. For each value of $\varepsilon$, in fact, we obtain a different estimate $\widehat{s}^{\varepsilon}$, with a different topological structure: for small values of $\varepsilon$, the topology of $\widehat{s}^{\varepsilon}$ is close to the one of the point--cloud itself. As $\varepsilon$ grows more and more points start to be connected, until eventually the corresponding $\widehat{s}^{\varepsilon}$ is homeomorphic to a point. 

\begin{figure}
\includegraphics[width=1\linewidth]{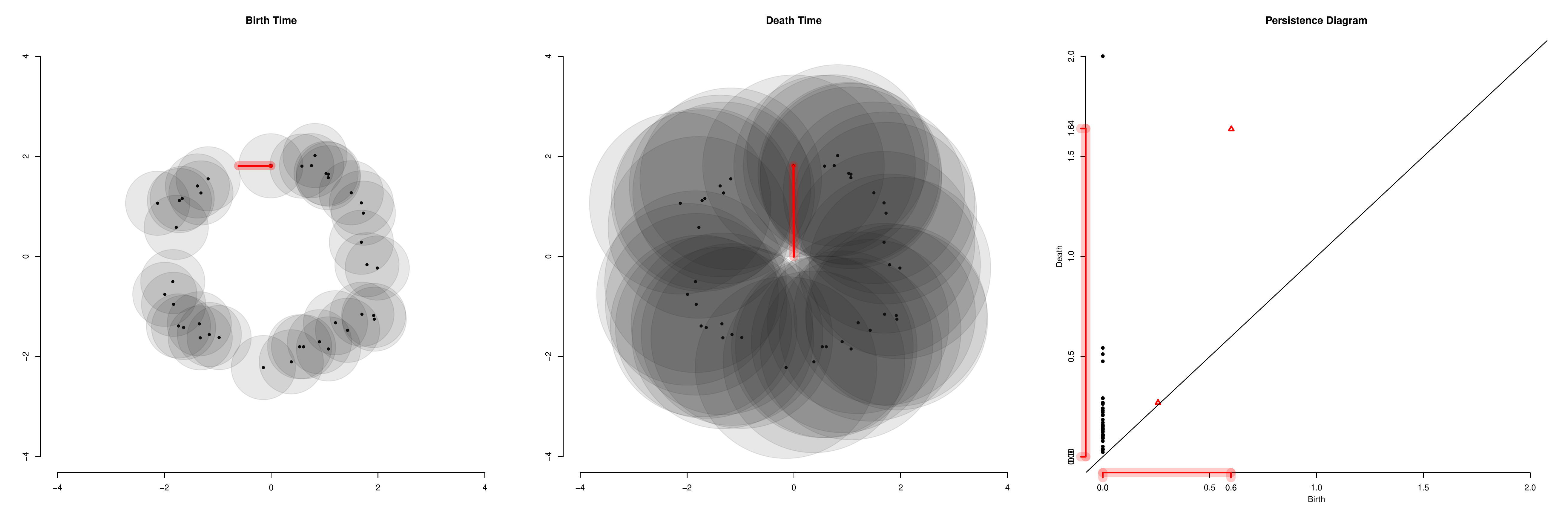}
\label{fig:image1}
\caption{From left to right: birth of the circle in the filtration, $\widehat{s}_{b}$, death of the circle $\widehat{s}_{d}$ and summarizing Persistence Diagram.  }
\end{figure}

The basic idea is that as $\varepsilon $ grows, different estimates $\widehat{s}^{(\varepsilon)}$ are related, so that if a feature is present in both we can say that it remains alive. Formally this corresponds to the notion of Persistent Homology, a multiscale version of Homology that allows to see how features appear and disappear at different scales. Values $\varepsilon_b, \: \varepsilon_d$ of $\varepsilon$ corresponding to when two components are connected for the first time (\emph{birth--step}) and when they are connected to some other larger component (\emph{death--step}) are the generators of a Persistent Homology Group (Figure \ref{fig:image1}).

\paragraph{Kernel Density estimators} The second way of recovering the topology of  $\mathcal{M}$ is based on the fact that superlevel set of a density function  $p$ can be topologically equivalent to the support of the distribution \cite{Fasy2014}. More formally, if the data are sampled from a distribution  $P $ supported on  $\mathcal{M}$, and if the density  $p $ of  $P $ is smooth and bounded away from  $0$, then there is an interval  $[\eta,\delta] $ such that the superlevel set  $p^{(\varepsilon)} =\{x\: | \: p(x)\geq \varepsilon\}$ is homotopic (i.e. topologically equivalent) to  $\mathcal{M} $, for  $\eta\leq \varepsilon \leq \delta $.

We do not know $p$, but we can approximate it with a kernel density estimator  $\widehat{p}_n $. A na{\"\i}ve way to estimate the topology of  $\mathcal{M} $ is thus to compute topological invariants of the superlevel set of the kernel density estimator  $\widehat{p}_n $:
\[
\widehat{p}^{(\varepsilon)}_n = \{x\: | \: \widehat{p}_n(x)\geq \varepsilon\}.
\]

The superlevel sets  $\widehat{p}^{(\varepsilon)}_n$, with $\varepsilon \in [0, \max \widehat{p}_n]$, form a decreasing filtration, which means that  $\widehat{p}^{(\varepsilon)}_n \subset \widehat{p}^{(\delta)}_n$ for all  $\delta\leq\varepsilon$. As in the case of distances, for each element in the filtration, i.e. for each value  $\varepsilon$, we obtain a different estimate $\widehat{p}^{(\varepsilon)}_n$, whose topology can be characterized  by its Homology Groups. Since in practice it is not possible to determine the interval  $[\eta,\delta]$ in which the topology of $\widehat{p}^{(\varepsilon)}_n$, is closest to that of $\mathcal{M}$, we analyse the evolution of the topology in the whole filtration. Persistent Homology allows to analyze how those Homology Groups change with $\varepsilon$.

\begin{figure}
\includegraphics[width=\linewidth]{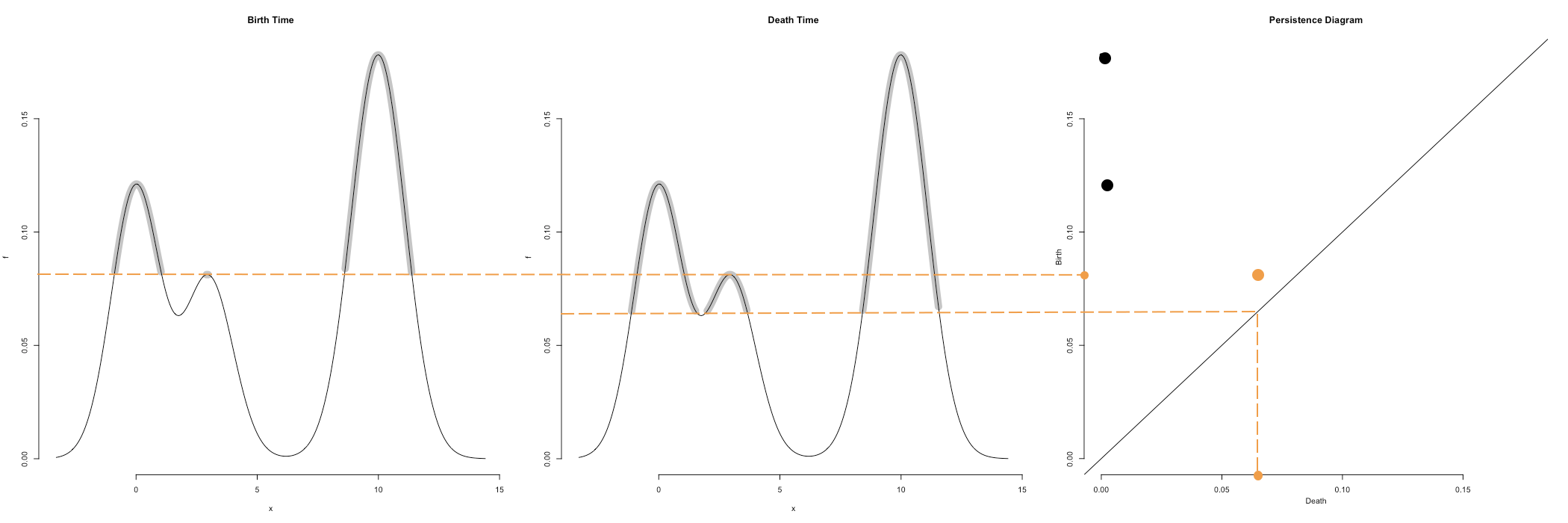}
\caption{From left to right: birth of the smallest peak in the filtration, $\widehat{p}^b _{n}$, death of the circle $\widehat{p}^{d}_n$ and summarizing Persistence Diagram.}
\label{fig:stolen}
\end{figure}

As can be seen from Figure \ref{fig:stolen}, connected components in the filtration  $\widehat{p}^{(\varepsilon)}_n$, can be though of as local maxima of  $\widehat{p}_n $, analogously, loops in  $\widehat{p}^{(\varepsilon)}_n$ represent circular structures in $\widehat{p}_n$ and so on. This is true for the distance function as well, although since the filtration is defined in terms of sublevel sets, connected components represent local minima instead.
In this sense, Persistent Homology can be considered a characterization of the whole function $f$, and extended to any arbitrary levelset filtration.

\subsection{Persistence Diagram} Persistent Homology Groups can be summarized by the \emph{Persistence Diagram}, a multiset $D = \{z_i = (b_i, d_i)\}_{i=1}^m$ whose generic element $(b_i, d_i)$ is the $i^{\tt th}$ generator of the Persistent Homology Group. Features with a long ``lifetime'' (or persistence $\text{pers} =  b-d$) are those which can be found at many different resolution of the filtration, and are informative of the topology of $\mathcal{M}$. Points that are close to the diagonal instead represent short--lived features, which may be only noisy artifacts and can be neglected.

The space of Persistence Diagrams  $\mathcal{D}$ is a metric space, when endowed with the \emph{Bottleneck distance}, which, given two Persistence Diagrams $D$ and $D'$, is defined as
\[
d_B (D, D') = \inf_{\gamma} \sup_{x\in D } \big\| x - \gamma(x) \big\|_{\infty},
\]
where the infimum is taken over all bijections  $\gamma : D \mapsto D'$.

The Bottleneck distance allows us to compare Persistence Diagrams and to define their most important property: \emph{stability} \cite{Chazal2012}.

\begin{theorem}[Stability]
Let  $f$ and  $g$ be two functions on a triangulable space $\mathbb{X}$ and let $D_{f}, D_{g}$ be the Persistence Diagram built on their respective sublevel (or superlevel) set filtrations, then 
\[
d_B(D_f,D_g) \leq \norm{f - g}_{\infty},
\] 
where $\norm{f}_{\infty} = \sup_x |f(x)| $ is the $L^{\infty}$--norm.
\label{th:stability}
\end{theorem}

In the special case of $f = d_\mathbb{X}$ and $g = d_\mathbb{Y}$ two distance functions defined on two point--clouds  $\mathbb{X}$ and $\mathbb{Y}$ respectively, the stability result can be written in a more easily interpretable way:
\[
d_B \left(D_{\mathbb{X}}, D_{\mathbb{Y}} \right) \leq 2 \, d_{H} \left( \mathbb{X}, \mathbb{Y} \right),
\]
where  $d_H(\mathbb{X},\mathbb{Y})$ is the \emph{Hausdorff} distance between two topological spaces  $\mathbb{X}$ and  $\mathbb{Y}$. Roughly speaking this means that if the two point clouds $\mathbb{X}$ and $\mathbb{Y}$ objects are close, their persistence diagrams will be as well, and can be interpreted in two ways:

\begin{itemize}
\item \textit{the persistence diagram is a topological signature:} stability reassures us that if two point-clouds $\mathbb{X,Y}$ are similar their Persistence Diagrams will be as well, and is therefore instrumental for using them in statistical tasks such as classification or clustering;

\item \textit{the persistence diagram is statistically consistent:} stability reassure us that if we are using a point--cloud $\mathbb{X}_n$ to estimate the topology of an unknown object $\mathbb{X}$, if $\mathbb{X}_n\rightarrow \mathbb{X}$ as $n\rightarrow \infty$, then $D_{\mathbb{X}_n}$ converges to $D_{\mathbb{X}}$ as well.
\end{itemize}

\section{Persistence Landscape}\label{persistence-landscape}

Persistence Diagrams are general metric objects, but several tools have been developed to convert them into functional objects, in order to work with more statistics-friendly spaces. The most famous transformations of the Persistence Diagram are the Persistence Landscape \cite{Bubenik2015} and the Persistence Silhouette \cite{Chazal2014a}, both
built by mapping each point  $z = (b, d)$ of a Persistence Diagram  $D$ to a piecewise linear function called the ``triangle'' function  $T_z$, which is defined as: 
\[
T_{z} (y) = (y-b+d)\mathbbm{1}_{[b-d,b]} (t) + (b+d-y)\mathbbm{1}_{(b, b+d]} (y).
\]
$\mathbbm{1}_A (x) $ is the standard indicator function: $\mathbbm{1}_A (x) = 1$ if  $x \in A$ and  $\mathbbm{1}_A (x) = 0$ otherwise. Informally a triangle function links each point of the diagram to the diagonal with segments parallel to the axes, and then rotates them of $45$ degrees.

The triangles  $T_z$ can be combined in many different ways. If we take their  $k\text{-}\!\max$, i.e.~the  $k^{\tt th}$ largest value in the set  $T_z(y)$, we obtain the $k^{\tt th}$ \emph{Persistence Landscape} 
\[
\lambda^k_{D}(y) = k\text{-}\!\max _{z\in D} T_z (y) \qquad k \in \mathbb{N}^+ .
\]
The Persistence Landscape $\lambda_{D}$ is the collection of functions $\{\lambda^k_{D}(y)\}_k$. If we take the weighted average of the functions  $T_z(y) $, we have the \emph{Power Weighted Silhouette} 
\[
\psi_p(t) = \frac{\sum_{z \in D} w_z ^p \, T_z(y)}{\sum_{z\in D} w_z ^p}.
\]

A Persistence Landscape  $\lambda_D$ is a representation of a Persistence Diagram  $D$ as a collection  $\{\lambda^1_D, \ldots, \lambda_D^K\}$ of piecewise linear functions, indexed by the order of the maximum to be considered in defining the landscape, $k$.

While the space of Persistence Diagrams  $\mathcal{D} $ is only a metric space, Persistence Landscapes are defined in a much richer Banach space $\mathcal{L} $, endowed with the following norm
\[
\norm{ \lambda_D }^p _p = \sum_k \norm{ \lambda^k }^p _p,
\] 
where $\norm{ \lambda^k }_p $ is the $L^p$--norm
\[
\norm{ \lambda^k } _p = \left(\int \lambda^k \text{d}\mu \right)^{1/p}.
\]

It is not possible to go back from Persistence Landscapes to Persistence Diagrams, meaning that there is a loss of information in going from Persistence Diagrams to Persistence Landscapes. However the Persistence Landscape is still informative, since stability still holds \cite{Bubenik2015}.

\begin{theorem} 
Let  $f,g$ be two functions on  $\mathbb{X}$ and let $D_f$ and $D_g$ be the Persistence Diagrams built from their superlevel (or sublevel) sets, then
\[
d_{\Lambda} \left( \lambda_{D_f}, \lambda_{D_g}\right) \leq \norm{f-g}_{\infty},
\] 
where $d_{\Lambda} \left( \lambda_{D_f}, \lambda_{D_g}\right) = \norm{ \lambda_{D_f} - \lambda_{D_g} }_{\infty}$ is the  $L^{\infty}$--distance in the space of Persistence Landscapes, $\mathcal{L}$. \label{th:landstability}
\end{theorem}

\begin{figure}
\includegraphics[width=1\linewidth]{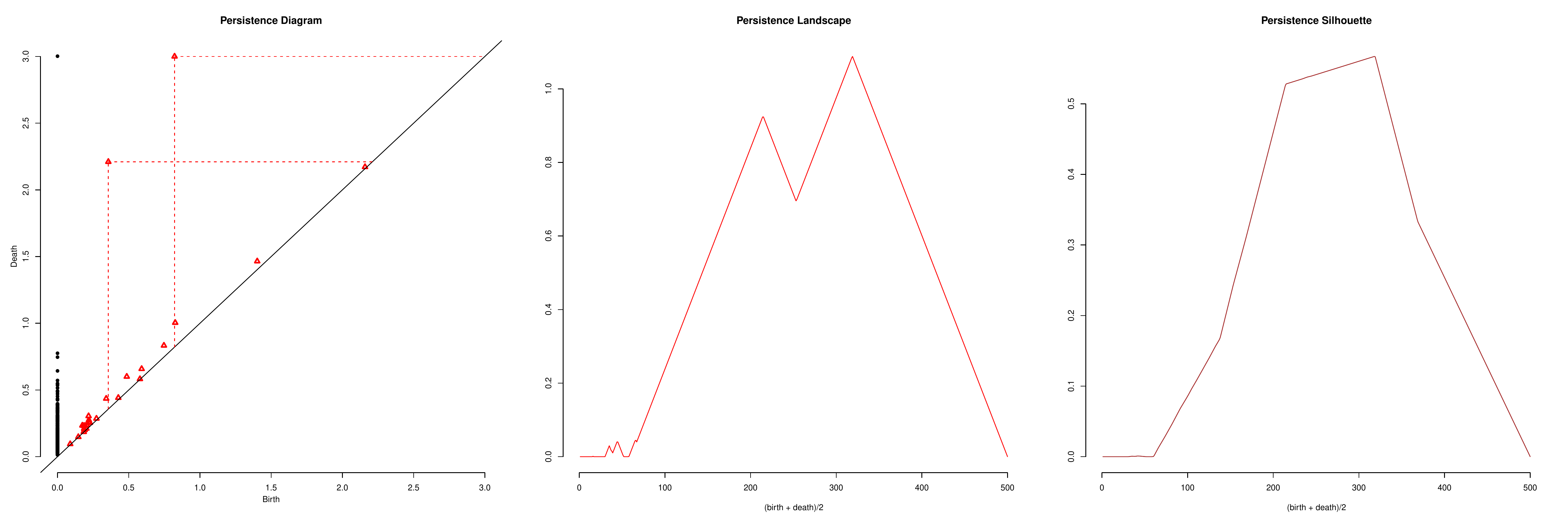} 
\caption{A Persistence Diagram (left) and its corresponding Persistence Landscape (center) and Persistence Silhouette (right).}
\end{figure}

Moreover the Persistence Landscape has a noticeable advantage with respect to Persistence Diagram, that is it is defined in a Banach Space and it can they can be considered as random variables, which is instrumental in statistical learning. 

\subsection{Probability in Banach Spaces / A modicum}\label{probability-in-banach-spaces.}

In order to better understand the desirable properties of topological summaries defined in a Banach space rather than in just a metric one, we quickly review the basic of Probability in Banach spaces; a more complete overview can be found in \cite{Ledoux2013}. Let  $\mathcal{B} $ be a real, separable Banach space with norm $\norm{ \cdot }$. Let $(\Omega, \mathcal{F}, \mathcal{B})$ be a probability space and let
\[
V:(\Omega, \mathcal{F}, \mathcal{B}) \mapsto \mathcal{B},
\] 
be a Borel random variable with values in  $\mathcal{B}$.

We call an element of  $\mathcal{B}$ the \emph{Pettis integral} of $V$ if $\mathbb{E}\big( f(V) \big) = f\big( \mathbb{E}(V) \big)$ for all $f \in \mathcal{B}^\star$, where $\mathcal{B}^\star$ is the space of continuous linear real--valued functions on  $\mathcal{B}$, i.e.~the topological dual space of $\mathcal{B}$. The \emph{Pettis integral} is the analogous of the expected value for a  $\mathcal{B}$--valued random variable. The following proposition gives us a sufficient condition for its existence.\\

\begin{proposition}
If  $\mathbb{E}\norm{V} < \infty$, then $V$ has a \emph{Pettis integral} and $\norm{\mathbb{E}(V)} \leq \mathbb{E}\norm{V}$.
\end{proposition}

Notice that $\norm{V}$ is a real valued random variable.

The Pettis integral can be used to define an extension of the Law of Large numbers for a $\mathcal{B}$--valued random variable. Recall that for a sequence $\{Y_n\}_n$ of $\mathcal{B}$--valued random variables:

\begin{itemize}
\item  $\{Y_n\}_n$ converges \emph{almost surely} to a $\mathcal{B}$--valued random variable  $Y$ if $\mathbb{P}(\lim_{n\rightarrow \infty} Y_n) = 1 $.

\item $\{Y_n\}_n$ converges \emph{weakly} to a $\mathcal{B}$--valued random variable $Y $ if $\lim_{n\rightarrow \infty} \mathbb{E}\big( \phi(Y_n) \big) = \mathbb{E}\big( \phi(Y) \big)$ for all bounded continuous functions $\phi: \mathcal{B} \mapsto \mathbb{R}$.
\end{itemize}

\begin{theorem}[Strong Law of Large Numbers] Let  $\{V_n\}_{n\in\mathbb{N}} $ be a sequence of independent copies of $V$  and, for a given $n$, let $S_n = V_1 + \cdots + V_n$,
\[
\frac{S_n}{n}\rightarrow \mathbb{E}(V)  \quad \text{almost surely} \iff \mathbb{E}\norm{V} < \infty.
\]
\label{th:LLN}
\end{theorem}
There is an extension of the Central Limit Theory as well, which states the convergence to a Gaussian random variable. In a Banach Space  $\mathcal{B}$, a random variable $G$ is said to be \emph{Gaussian} if for each $f \in \mathcal{B}^\star$, $f(G)$ is a real valued Gaussian random variable with $0$ mean. The covariance structure of a $\mathcal{B}$--valued random variable, which fully characterize a Gaussian Random Variable in a Banach Space, is given by 
\[
\mathbb{E}\big[ \big( f(V)-\mathbb{E}[f(V)] \big) \cdot \big( g(V)-\mathbb{E}[g(V)] \big) \big],
\]
where $f,g \in \mathcal{B}^\star$.

\begin{theorem}[Central Limit Theorem] Assume  $\mathcal{B} $ has type  $2 $. If  $\mathbb{E}(V)=0 $ and  $\mathbb{E}(\norm{V}^2)<\infty $ then  $\frac{S_n}{\sqrt{n}} $ converges weakly to a Gaussian random variable  $G(V) $ with the same covariance structure as  $V $.
\label{th:CLT}
\end{theorem}

The extension of these two result to the case of Persistence Landscapes
is immediate.

\section{The Persistence Flamelets}\label{the-landscape-flow}

Persistence Diagrams and Persistence Landscapes gives us a full characterization of a function $f$ in terms of the topology of its sub- (or super-)levelset filtration. Inspired by \emph{scale space theory} we now investigate the evolution in the topology of $f$ when we allow it to continuously change with respect to some scale parameter $\sigma$; that is, when the focus of the analysis becomes a family $\mathcal{F} = \{f_\sigma, \; \sigma \in S\}$, where $S$ is some bounded set\footnote{For the sake of simplicity we will assume $\sigma\in [0,1]$, as every bounded set can be rescaled to $[0,1]$.}. Our goal is to simultaneously summarize the topology at each resolution $f_\sigma$ and how it changes with $\sigma$.

The most intuitive way of encoding a scale parameter into the \texttt{TDA} framework is to consider as a function of $\sigma$ the family of Persistence Diagrams $\mathbb{D} = \{D_\sigma, \; \sigma \in [0,1] \} $ corresponding to $\mathcal{F}$.  $\mathbb{D} $ is known as Persistence Vineyards  \cite{Cohen-Steiner2006} and is a stable and continuous \cite{Morozov2008} representation of the topology of the whole  $\mathcal{F} $. However, Persistence Vineyards share all the drawbacks and limitations of Persistence Diagrams, more specifically they lack a unique average and a measure of variability for a group of them \cite{Turner2014}. Moreover, it is not yet clear whether or not it is possible to explicitly define a probability distribution on the space of Persistence Diagrams (and consequently on the space of Persistence Vineyards), which severely limits their use in statistical inference \cite{Mileyko2011}. 

We thus introduce a new representation, based on the Persistence Landscape, that overcomes most of these issues. It is worth noticing that although in the following we focus on Persistence Landscapes, the same results hold for Silhouettes as well. Our basic idea is to consider the Persistence Landscapes $\lambda_{D_\sigma}$ corresponding to the family $\mathcal{F} = \{f_\sigma, \; \sigma \in [0,1]\} $ as a function of the scale parameter  $\sigma $. Visually we can think of such function as a ``flow'' of landscapes, one for each resolution, smoothly moving and resembling a tiny fire (see, for example, Figure \ref{fig:quakesSiZer}). \\

\begin{definition}[Persistence Flamelets]
Given a collection of Persistence Diagrams $D_\sigma $, continuously indexed by some parameter  $\sigma \in [0,1]$, and  $k \in \mathbb{N}^+ $, we define the  $k^{\tt th}$ \emph{Persistence Flamelets} as the function
\[
\Lambda^k(\sigma, y) = \lambda^k_{D_\sigma}( y) \qquad \quad \forall \, \sigma \in [0,1],\: y \in \mathbb{R},\: k\in \mathbb{N}^+.
\]
As the Landscape itself, the \emph{Persistence Flamelets}  $\Lambda$ is also a collection  $\Lambda = \{\Lambda^{(k)}, \; k \in \mathbb{N}^+ \}$
indexed by the order of the $\max$ we consider.
\end{definition}

The theoretical reassurance that the Persistence Flamelets is a meaningful topological summary is its stability, which we will prove in the following. Before doing so, however, we need to introduce a notion of \emph{proximity} between Persistence Flamelets.\\

\begin{definition}[Integrated Landscape distance] Let  $\mathbb{D}=\{D_\sigma, \; \sigma \in [0,1]\},\; \mathbb{E}=\{E_\sigma, \; \sigma \in [0,1]\}$ two Persistence Vineyards and  $\Lambda_{\mathbb{D}}, \Lambda_{\mathbb{E}} $ the corresponding Persistence Flamelets. We define the \emph{Integrated Landscape distance} between  $\Lambda_{\mathbb{D}} $ and  $\Lambda_{\mathbb{E}}$ as
\[
I_{\Lambda}( \Lambda_{\mathbb{D}}, \Lambda_{\mathbb{E}}) = \int_0^1 \!\! d_{\Lambda} (\lambda_{D_\sigma}, \lambda_{E_\sigma}) \, \text{d}\sigma.
\]
\end{definition}

\begin{theorem} Let  $\mathbb{D}=\{D_\sigma, \; \sigma \in [0,1]\},\; \mathbb{E}=\{E_\sigma, \; \sigma \in [0,1]\}$ two Persistence Vineyards and $\Lambda_{\mathbb{D}}, \Lambda_{\mathbb{E}}$ the corresponding Persistence Flamelets, then:

\begin{enumerate}
\def\labelenumi{\arabic{enumi}.}
\item $\Lambda_{\mathbb{D}}$ and  $\Lambda_{\mathbb{E}}$ are continuous with respect to the Bottleneck distance;

\item $I_{\Lambda}(\Lambda_{\mathbb{D}}, \Lambda_{\mathbb{E}}) \leq I_B (\mathbb{D}, \mathbb{E})$
\end{enumerate}

where $I_B (\mathbb{D}, \mathbb{E}) = \int_0 ^1 d_{B} (D_\sigma, E_\sigma) \, \text{d}t$ is the Integrated Bottleneck distance for Persistence Vineyards as defined in \cite{Munch2013}.
\label{th:flowstability}
\end{theorem}

The proof is a direct consequence of the Stability Theorem for Persistence Landscapes (Theorem \ref{th:landstability}) and the continuity of Persistence Vineyards, in fact:

\begin{enumerate}
\item  For a fixed  $\sigma $, consider  $D_\sigma $ and $D_{\sigma+\varepsilon} $ (same applies for  $\mathbb{E} $). By \ref{th:landstability} and the continuity of  $\mathbb{D} $ we have
\[
0 \leq \lim_{\varepsilon \rightarrow 0} d_{\Lambda} \left( \lambda_{D_{\sigma}}, \lambda_{D_{\sigma+\varepsilon}}\right)  \leq \lim_{\varepsilon \rightarrow 0} d_{B} \left(D_\sigma, D_{\sigma+\varepsilon}\right) = 0.
\]
\item Since for a fixed  $\sigma $ we have, by Theorem \ref{th:landstability} we have
\[
d_{\Lambda} \left( \lambda_{D_{\sigma}}, \lambda_{E_{\sigma}}\right)  \leq  d_{B} \left( D_{\sigma}, E_{\sigma} \right)
\]
integrating both terms is enough to prove the result.
\end{enumerate}

The Persistence Flamelets is also a random variable defined in a Banach space. In analogy with what \cite{Bubenik2015} has done for Persistence Landscapes, we define a norm for Persistence Flamelets, more specifically
\[
\norm{\Lambda}^p _p = \int_0 ^1 \sum_{k} \norm{\lambda^{k}(t)}^p_p \text{d}t
\]
Then following \cite{Ledoux2013}, we can extend the Law of
Large Numbers and the Central Limit Theorem to this new object.\\

\begin{corollary}[Strong Law of Large Numbers] Let $\{\Lambda_n\}_{n\in\mathbb{N}}$ be a sequence of independent copies of $\Lambda$ and, for a given $n$, let $S_n = \Lambda_1+\dots+\Lambda_n$, where the sum is defined pointwise.
\[
\frac{S_n}{n}\rightarrow \mathbb{E}(\Lambda)  \quad \text{almost surely} \iff \mathbb{E}\norm{\Lambda} < \infty.
\]
\end{corollary}

\begin{corollary}[Central Limit Theorem] Assume $\mathcal{B}$ has type $2$. If $\mathbb{E}(V) = 0$ and $\mathbb{E}(\norm{\Lambda}^2) < \infty$ then  $\frac{S_n}{\sqrt{n}}$ converges weakly to a Gaussian random variable $G(\Lambda)$ with the same covariance structure as $\Lambda$.
\end{corollary}

Proofs directly follow from Theorem \ref{th:LLN} and Theorem \ref{th:CLT}.

\subsection{Some intuition / EEG Dynamic Point--Clouds} \label{dynamic-point-clouds---eeg-data}

A short example will clarify when this object, until now very abstract, may be encountered and fruitfully used. The easiest way to understand the need for Multiscale Persistent Homology is to consider time as a scale parameter \cite{Munch2015, Munch2013}; the Persistence Flamelets allows for a characterization of the dynamic process $ \mathcal{F} = \{P_t, \; {t\in [0,1]}\}$ in terms of its topology.  

For each time $t$ we observe a sample  $\mathbb{X}(t) = \{X_1(t), \dots, X_k(t) \}$ drawn from $P_t$; the trace of the sample in the time interval $\{\mathbb{X}(t), \; t \in [0,1]\}$  is usually called a \emph{Dynamic Point Cloud}. The Persistence Flamelets allows us to simultaneously study the shape of $\mathbb{X}(t)$ and how it evolves with $t$, giving us a new type of insights on high dimensional time series. %This setting has received much attention in the context of \texttt{TDA} \cite{}, \cite{}, as it is often difficult to characterize relationships between 

In the special case of dynamic point--clouds, the stability result of Theorem \ref{th:flowstability} can be restated as follows.

\begin{corollary}
Let  $\{\mathbb{X}(t),\mathbb{Y}(t)\}$ with $t\in (0,1)$ two continuous dynamic point clouds, $\Lambda_{\mathbb{X}}$ and  $\Lambda_{\mathbb{Y}}$ their corresponding Persistence Flamelets, then:
\[
I_{\Lambda}(\Lambda_{\mathbb{X}}, \Lambda_{\mathbb{Y}}) \leq I_H (\mathbb{X}, \mathbb{Y}),
\]
where $I_H (\mathbb{X}, \mathbb{Y}) =\int_0 ^1 d_{H} \big( \mathbb{X}(t), \mathbb{Y}(t)\big) \text{d}t$ is the Integrated Hausdorff distance for dynamic point--clouds, as defined in \cite{Munch2013}.
\end{corollary}

\begin{figure}
\includegraphics[width=.5\textwidth]{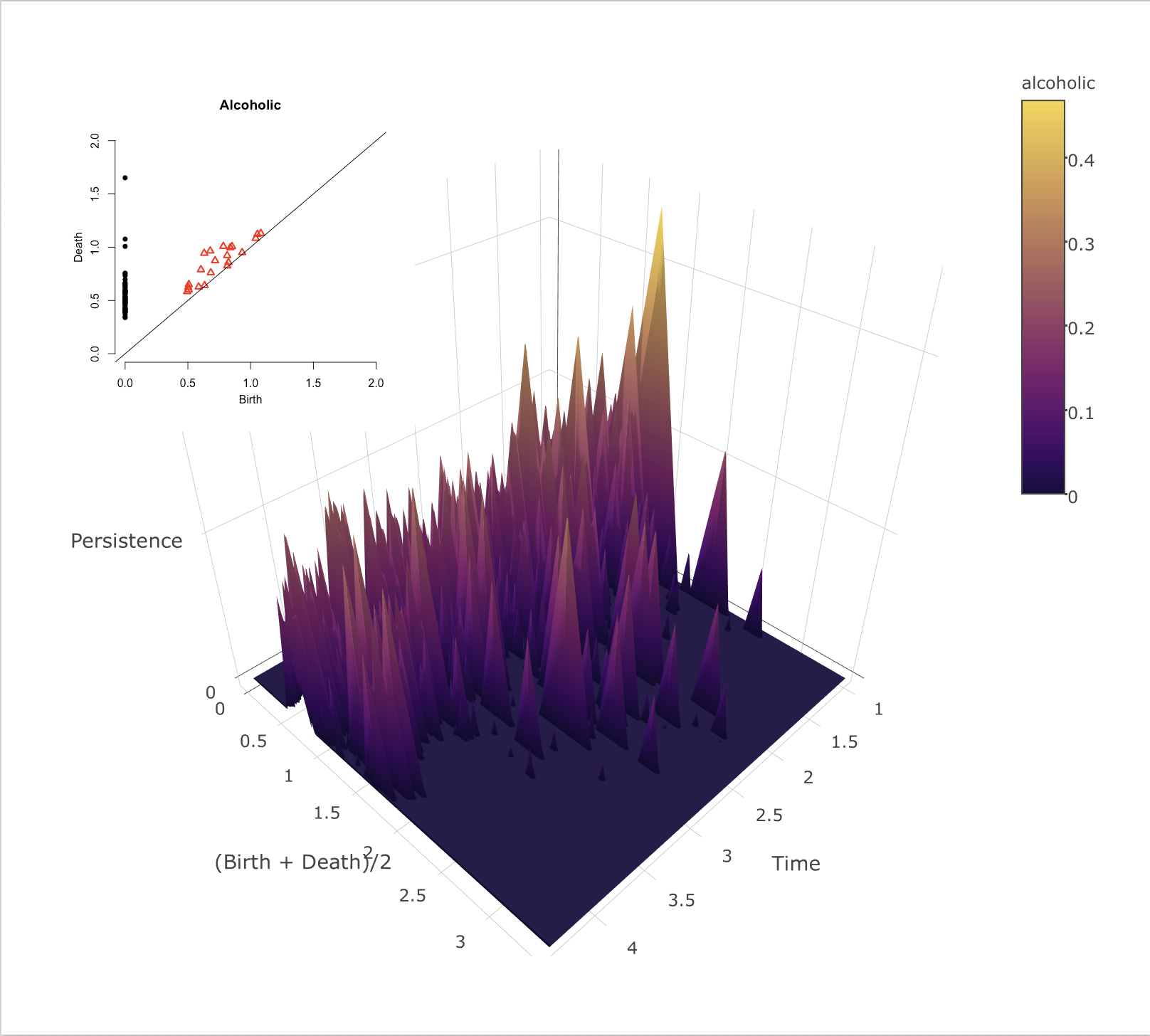} 
\includegraphics[width=.5\textwidth]{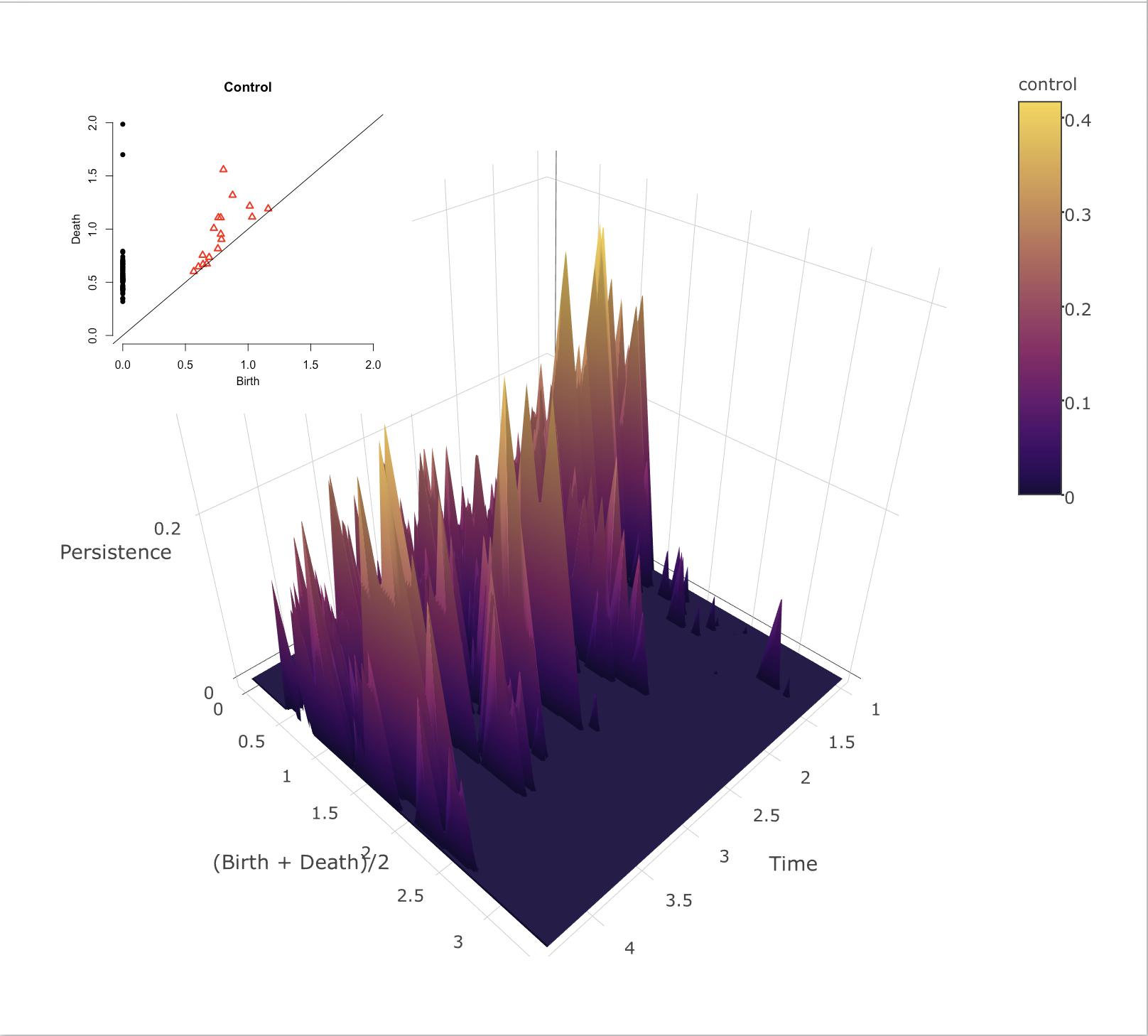} 
\caption{Persistence Flamelets of Dimension $1$ for the EEG data of one alcoholic (left) and one control (right) subject.}
\label{fig:comp2}       
\end{figure}

Figure \ref{fig:comp2} shows two Persistence Flamelets built from electroencephalography (EEG) tracks, freely available on the \href{https://archive.ics.uci.edu/ml/datasets/eeg+database}{\tt UCI Machine Learning Repository}. EEG are electric impulses recorded at a very high frequency ($256$ \emph{Hz}) through multiple electrodes ($64$ in this study), located in different areas of the skull. This kind of data fits perfectly in our framework; at each time  $t$, connected components and loops represent area of the brain that share the same behavior, which is relevant information per se, but it is also important to assess whether or not these connection persist in time.

We compare the EEGs of one alcoholic (left) and one control (right) patient, both subject to the same stimulus. For each of them we have $5$ trials of  $1$ second; since EEG are typically very noisy, we average them across repetition before computing their topological summaries. Persistent Homology is computed using the \texttt{R} package \texttt{TDA} \cite{Fasy2017}. Results here shown refer to dimension $1$ features (loops) but similar conclusions could be drawn from the dimension $0$ features as well.

The Persistence Flamelets highlights differences in the brain's behavior of the two individuals. The signal from the control patient, in fact, is strongly characterized by a few persistent features. In the alcoholic patient instead there is less structure; there seems to be more features than in the control patient, but they all have a smaller persistence, and could therefore be interpreted as noise.

\section{Scale Space Methods in Data Smoothing}\label{kernel-density-bandwidth-exploration}

Although it may be useful in multiple settings, from time series to spatial modelling, where the scale may be given by time and space, %assuming that we take the resolution to be our scale parameter for example, 
the Persistence Flamelets is particularly relevant in the context of smoothing, where it allows to summarize and evaluate the evolution of the whole smoothing process. 

Roughly speaking, data smoothing is a family of methods aimed at recovering some structure in the data. Depending on their scale, however, smoothing methods may enhance noise or neglect relevant features, so that it is crucial to understand the impact of the smoothing level on the estimates.

The problem of assessing whether or not a features in a smooth is worth considering has been tackled in two very different ways:
\begin{itemize}
\item \textit{Selection}. This is the standard approach and consists of picking an optimal level of smoothing, typically using some resampling method \cite{Jones1996, Rudemo1982}. Features are then taken to be meaningful if they appear in the optimally smoothed data, and noise otherwise. 

\item \textit{Exploration}. This is the so--called scale space approach \cite{Lindeberg1994}, which rather than focusing on one level explores all of them, so that all features may be meaningful, but at different resolutions.
\end{itemize}

We claim that the Persistence Flamelets can be of use in both approaches. Since the Persistence Flamelets is multiscale by definition, it is natural to exploit it in the scale--space framework. We will show, however, that it also plays a role in the context of selection, and that it can be used to choose a ``topologically--aware'' bandwidth.

\subsection{Kernel Density Bandwidth Exploration}

% Among all the smoothing methods, the one for which the 
The smoothing method for which the problem of assessing the level of smoothness has undergone the most intensive study is Kernel Density Estimation \cite{Scott2015}. Part of the motivation behind the interest in this particular procedure is that the features affected by the smoothing process, typically local peaks (or, in topological terms, $0^{\tt th}$ dimensional Homology Groups), are meaningful in statistics, having a particularly relevant interpretation: local modes of a density and their basin of attraction represent are in fact one way of defining clusters \cite{Ester1996,Comaniciu2002}.

Given a sample $\{X_1 \ldots, X_n\}$, drawn from some smooth density $p$, a Kernel Density Estimator $\widehat{p}_h$ is defined as
\[ 
\widehat{p}_h(x) = \frac{1}{n}\sum_{i=1} ^n K_h(x - X_i),
\] 
where $K_h(x - y) = \frac{1}{h} K(\frac{x - y}{h})$ is a scaled kernel, $h$ is the bandwidth parameter and $K(\cdot)$, the kernel, is a non-negative, symmetric function that integrates to $1$.

While any kernel function $K(\cdot)$ may be used without compromising the performance of the estimator, the bandwidth parameter represent the level of smoothing and needs to be finely tuned. In the scale-space approach, given some bounded range of bandwidths $H \subset \mathbb{R}^+$, all the estimators $\widehat{p}_h$ are simultaneously considered, so that the object of interest becomes the family of smooths $\mathcal{F} = \{\widehat{p}_h \,:\, h\in H\}$. Since $K_h$ is continuous with respect to $h$ by definition, it is immediate to see that the Persistence Flamelets can be used to investigate and characterize $\mathcal{F}$.   %kernel  can be arbitrarily chosen, the only requirement being its continuity with respect to the bandwidth parameter $h\in \mathbb{R}^+$. We stress the fact that the choice of the kernel $K$  does not affect the estimate as much as the choice of the bandwidth \cite{}. 

% \texttt{SiZer}: The idea is to highlight significant features, such as bumps, by displaying where (with respect to both location and scale) the curve significantly increases and decreases. Note that significant bumps will be at zero crossings of the derivative between regions of significant increase and decrease. The name "\texttt{SiZer}" is a shortening of "SIgnificant ZERo crossings of derivatives." The color scheme is blue (red) in locations where the curve is significantly increasing (decreasing), and the intermediate color of purple is used where the curve cannot be concluded to be either decreasing or increasing. Here the term "loca- tion" is used in the scale-space sense of both "x-location" and "bandwidth location." Gray is used to indicate regions where the data are too sparse to make statements about significance, because there are not enough points in each window, as defined precisely in Section 3.

The first attempt at investigating the relation between the bandwidth of a kernel density estimator and its topology \texttt{SiZer} \cite{Chaudhuri1999}. Roughly speaking, given a sample $\{X_1, \ldots, X_n\}$ drawn from a univariate density $p$, \texttt{SiZer} (SIgnificant ZERo crossings of derivatives) is a map showing where in space, $x$, and scale, $h$, the kernel density estimator $\widehat{p}_h(x)$ is significantly increasing or decreasing. Since local peaks of a curve can be thought of as points where its derivative changes sign, the basic idea of \texttt{SiZer} is assess where this change happens, by testing whether the sign of the derivative $\widehat{p}'_h(x)$ for each couple of values $(x,h)$ is positive or negative. Values $(x,h)$ corresponding to significantly positive derivatives are shown in red and significantly negative are shown in blue, as in Figure \ref{fig:quakesSiZer}. 

\texttt{SiZer} is intrinsically $1$--dimensional and even though it has been extended to $2$--dimensional densities, especially in the context of image analysis, \cite{Godtliebsen2004} the features it hunts for are always and only local modes. The Persistence Flamelets provides a further extension in two different directions: 

%The Persistence Flamelets extends the idea of \texttt{SiZer} to arbitrarily high dimension, in two different ways:

\begin{itemize}
\item it can be used to investigate topological features of any dimension, rather than only feature of dimension $0$, i.e. local peaks;

\item it does not depend on the dimension of the data and can thus be used to investigate kernel densities for very high dimensional data.
\end{itemize}

Finally, even though, with respect to \texttt{SiZer}, the Persistence Flamelets lacks of statistical testing to asses the significance of each peak, it provides a measure of the relevance of each feature, its persistence.

% however, by features one typically refers to bumps. We claim that the Persistence Flamelets can extend this framework so to take into account also higher dimensional and more complex structure. The Persistence Flamelets allows for a characterization of the features at different resolution and is therefore instrumental in understanding the effect\ldots{} The main advantage of the Persistence Flamelets is that while the Although what we state in the following is true for any smoothing methods, as long as they

\begin{figure}
\centering
\includegraphics[width=.49\linewidth]{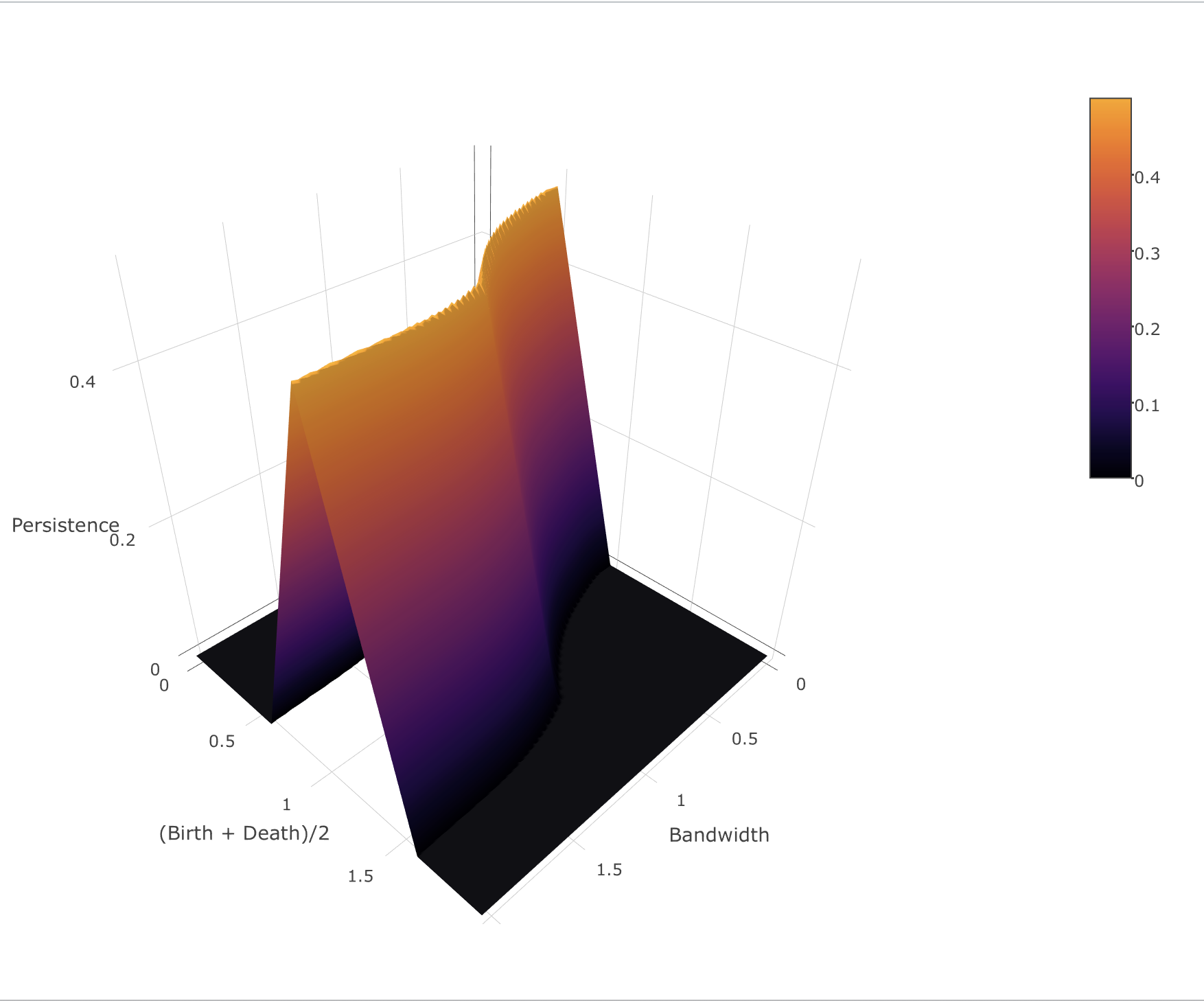} 
\includegraphics[width=.49\linewidth]{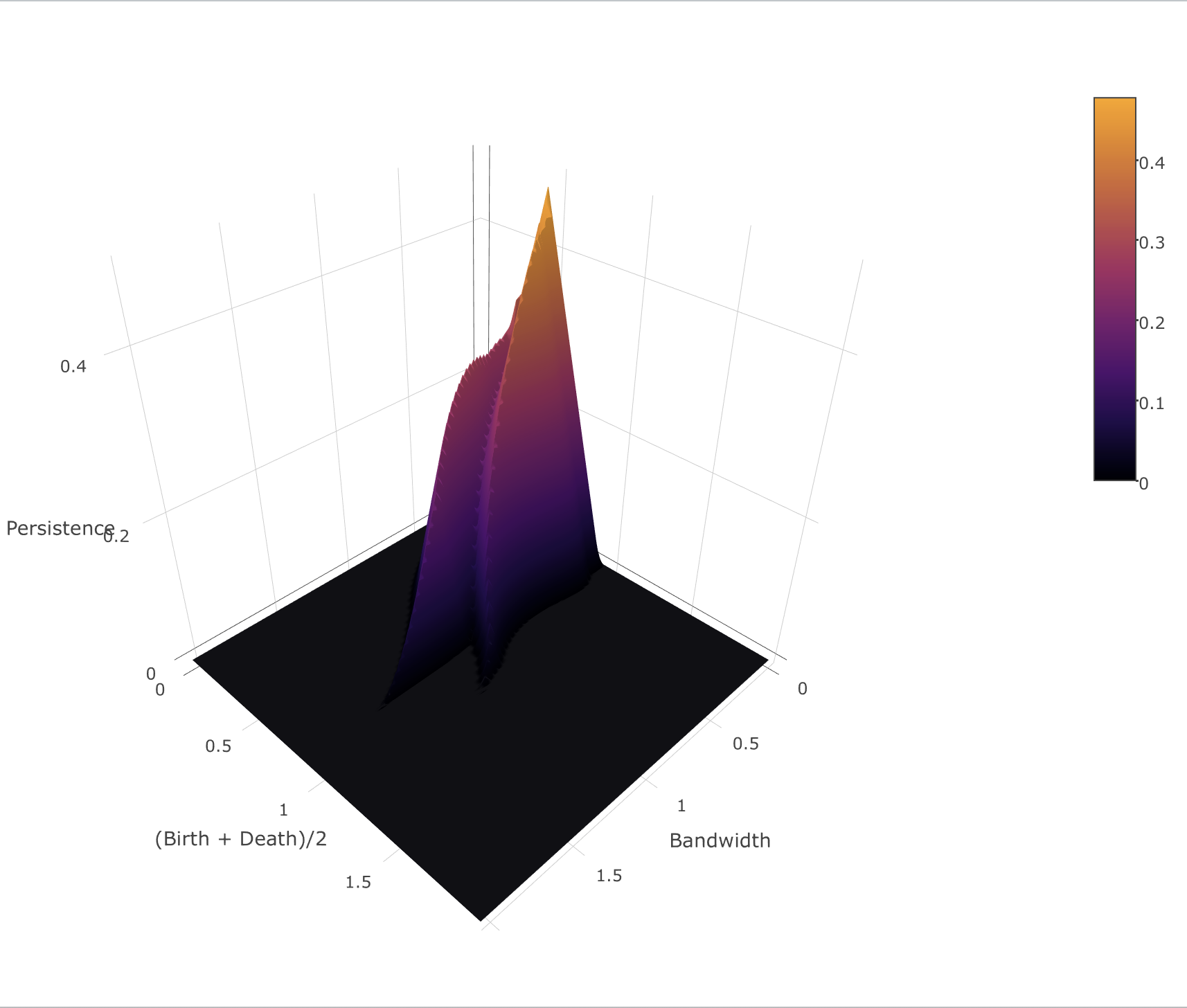} 
\caption{$1^{\tt st}$ (left) and $2^{\tt nd}$ (right) Persistence Flamelets of dimension $0$.}
\label{fig:fooquakes}
\end{figure}

% At each scale the bandwidth  $h $ affects the topology of the kernel
% density estimator. When  $h $ is small in fact, data are undersmoothed
% while when  $h $ is too large the opposite happens. 
% The connection between the bandwidth of the kde and its topology is an open and relevant question in statistics as the connected components represent local modes and their basin of attraction , as the superlevel set are important quantities in statistics. It is enough to think that connected components, represent local modes and their basin of attractions, which is one way to define clusters. connected components, r these quantities have a special meaning in statistics.

%The first attempt at analyzing the relation between the topology of a kde and its topology relation is Marron \& friend's Chaudhuri and Marron (1999) SiZer. {[}\textbf{review of SiZer}{]}.

%Since there are very many methods to pick a bandwidth a tool which does not even does that may seem useless, however, in the case of densities concentrated around lower dimensional objects, standard methods such as cross validation fail.

\subsection{Applications \& Comparisons}

We now show two real--data applications. In the first univariate one we quickly compare the Persistence Flamelets with \texttt{SiZer} and show that, when both are available they yield similar insights. The second is a bivariate example, which motivates investigating higher dimensional features and highlights the potential of the Persistence Flamelets when other tools are not available. 

\paragraph{Eartquakes I / Depth} In our first example we consider a classical dataset in kernel density estimation, the depth of the $512$ earthquakes beneath the Mt. St. Helens volcano in the months before the eruption of  $1982$ \cite{Scott2015}. Figure \ref{fig:fooquakes} shows the $1^{\tt st}$ and the $2^{\tt nd}$ Persistence Flamelets for the $0$ dimensional topological feature of the density estimator $\widehat{p}$ built with the Gaussian Kernel: 
\[
\widehat{p}_h(x) = \frac{1}{n}	 \sum_{i=1} ^n K_h(x - X_i) = \frac{1}{n}	\sum_{i=1}^n  \frac{1}{ \sqrt{2\pi h} } \exp\left\{\frac{1}{2h}(x -X_i)^2  \right\}.
\]
The $1^{\tt st}$ Persistence Flamelets consists of only one peak, representing the global maximum, which, as we can expect, always persists. This is not very informative, and when analyzing dimension  $0$ topological features, it is thus advisable to consider $2^{\tt nd}$ Persistence Flamelets, which represents the most relevant local peaks. 

\begin{figure}
\centering
\includegraphics[width=1\linewidth]{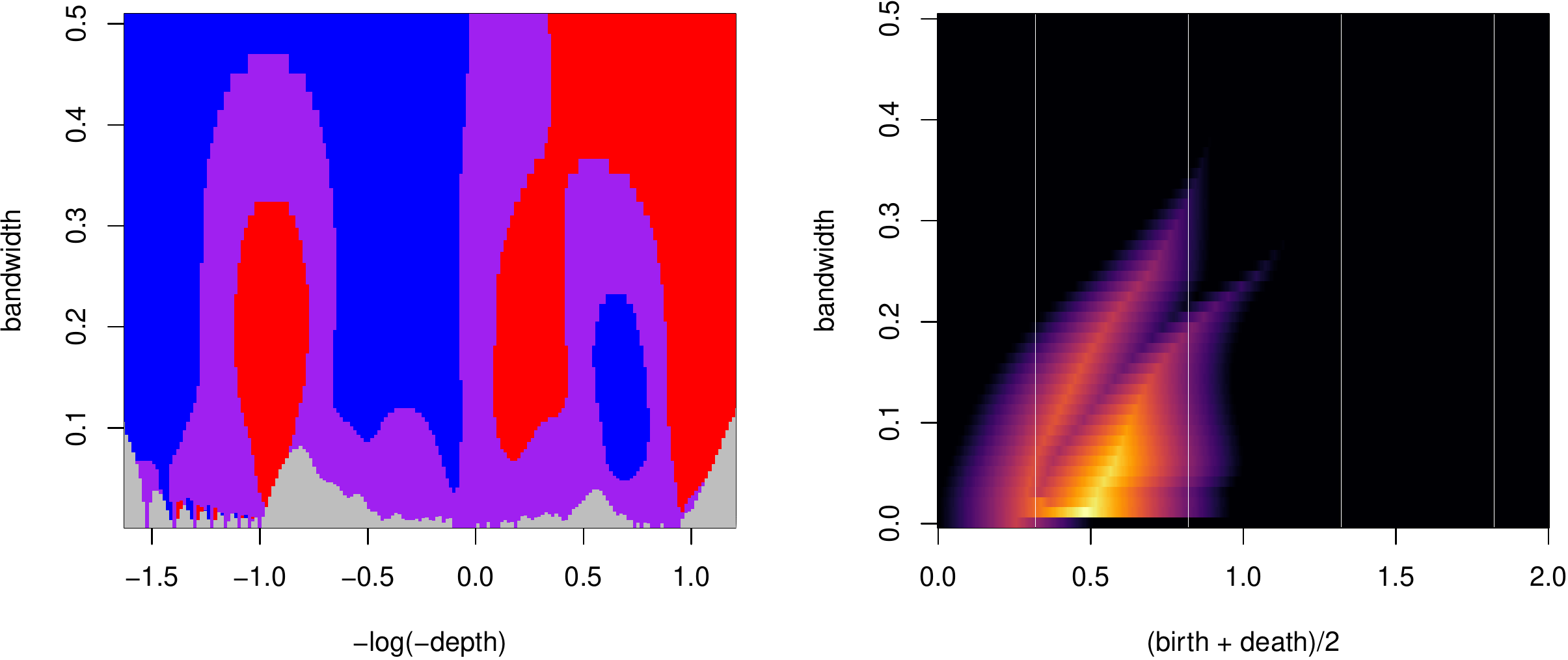} 
\caption{\texttt{SiZer}, the $1^{\tt st}$ and $2^{\tt nd}$ Persistence Flamelets of dimension $0$. In order to facilitate the comparison with \texttt{SiZer}, the Persistence Flamelets is projected and represented as a matrix.}
\label{fig:quakesSiZer}
\end{figure}

In this case we can see that the two peaks appearing in the $2^{\tt nd}$ Persistence Flamelets correspond to the two points in the diagram (which in turn correspond to the two bumps we can see in the KDE in Figure \ref{fig:quakesSiZer}). As we can see from Figure \ref{fig:fooquakes}, the $2^{\tt nd}$ Persistence Flamelets behaves differently than $1^{\tt st}$ Persistence Flamelets; when the bandwidth grows in fact, the two secondary peaks are smoothed away. 

Figure \ref{fig:quakesSiZer} shows the comparison with \texttt{SiZer}, and it is easy to see that the two approaches lead to very similar conclusions. The three peaks appear for  $h = 0.05$, then one of them disappear at around $h = 0.25$, one other around  $h = 0.35$ and, the last one always survives (in the given range of bandwidths).

\paragraph{Earthquakes II / Locations}\label{toy-example---earthquakes}

\begin{figure}
\centering
\includegraphics[width=1\linewidth]{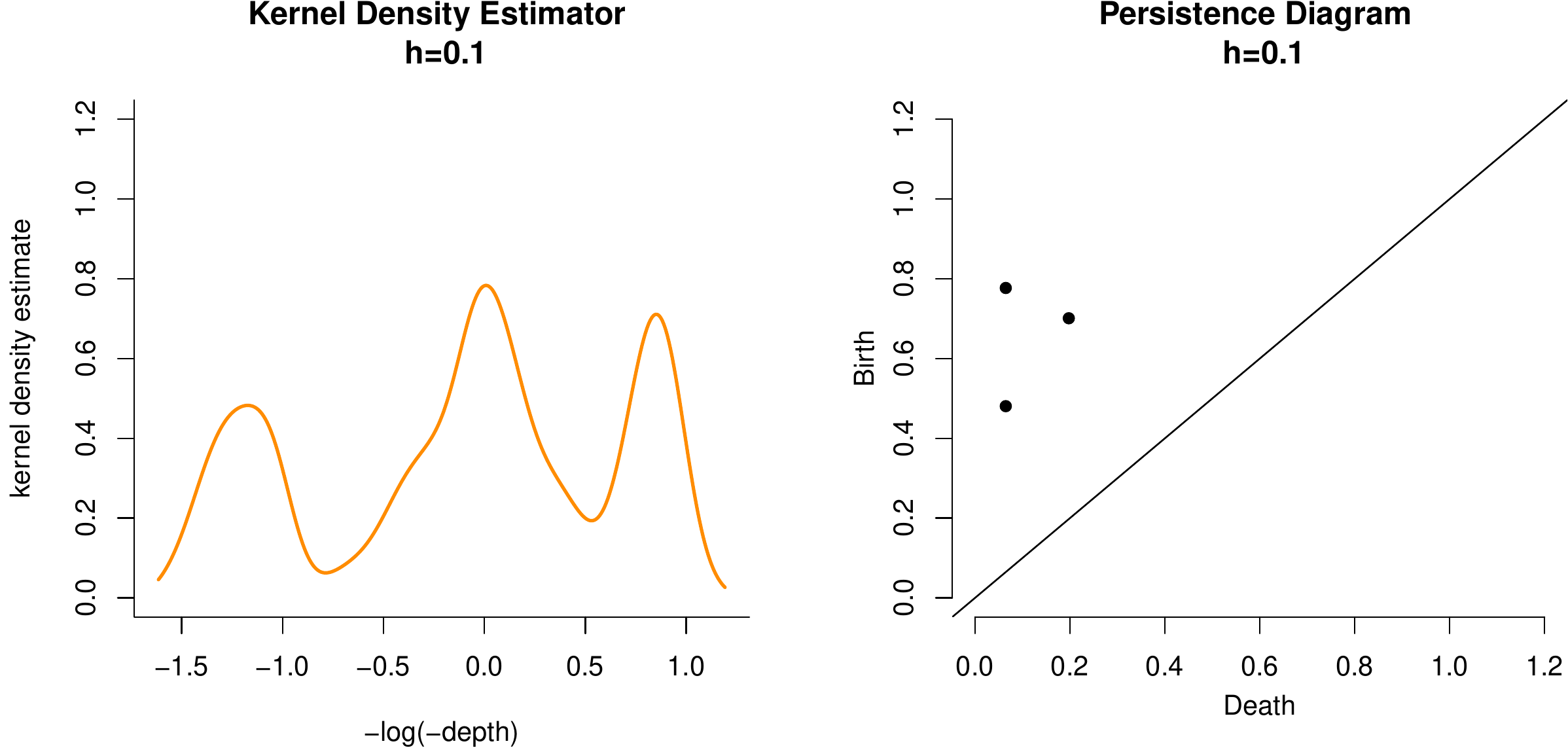} 
\caption{From left to right: Kernel Density Estimator of the Mt. St. Helens dept data (with $h = 0.1$) and corresponding Persistence Diagram.}
\end{figure}

For our second example we consider earthquake data coming from the \href{https://earthquake.usgs.gov/earthquakes/search/}{\tt USG catalog}. Our sample consists of the locations, expressed in latitude and longitude, of $6500$ events with magnitude higher than $5$, taking place between June $2013$ and June $2017$. The $2$--dimensional density $p$ generating the data $\{\bm{X}_1, \ldots, \bm{X}_n\}$ can still be estimated using the kernel density estimator with a Gaussian Kernel:

\begin{eqnarray*}
\widehat{p}(\bm{x}) & = & \frac{1}{n}	 \sum_{i=1} ^n K_{\bm{H}}(\bm{x} -\bm{ X}_i) \\
& = &  \frac{1}{n} 	 \sum_{i=1} ^n \frac{1}{2 \pi |\bm{H}|^{1/2}} \exp\left\{-\frac{1}{2} (\bm{x}-\bm{X}_i)^t  \bm{H}^{-1} (\bm{x}-\bm{X}_i)\right\}.
\end{eqnarray*}

Notice that in the multivariate case, the bandwidth is not a scalar but rather a matrix $\bm{H}$, however we chose an isotropic Gaussian Kernel, which corresponds to imposing a spherical structure to the covariance matrix 
\[ 
\bm{H} = h  
    \begin{pmatrix}
    1 & 0           \\
    0 & 1
    \end{pmatrix}, \qquad h\in \mathbb{R}^+,
\]
so that the kernel density estimator expression can be simplified as follows: 
\[
\widehat{p}(\bm{x}) = \frac{1}{n}	\sum_{i=1}^n  \frac{1}{ 2\pi h }	\exp\left\{-\frac{1}{2h^2}(\bm{x}-\bm{X}_i)^t (\bm{x}-\bm{X}_i) \right\}.
\]
Earthquakes are concentrated around circular structures, also known as \emph{plates}. According to Plate Tectonics, in fact, the Earth's lithosphere is broken into $7$ main plates, plus a number of minor ones. Since earthquakes are caused by the movements of neighboring plates, the density $p$ naturally inherits the Earth's plates structure. In terms of topology, plates can be thought of loops, or dimension $1$ Homology Groups. 

\begin{figure}
\includegraphics[width=.55\linewidth]{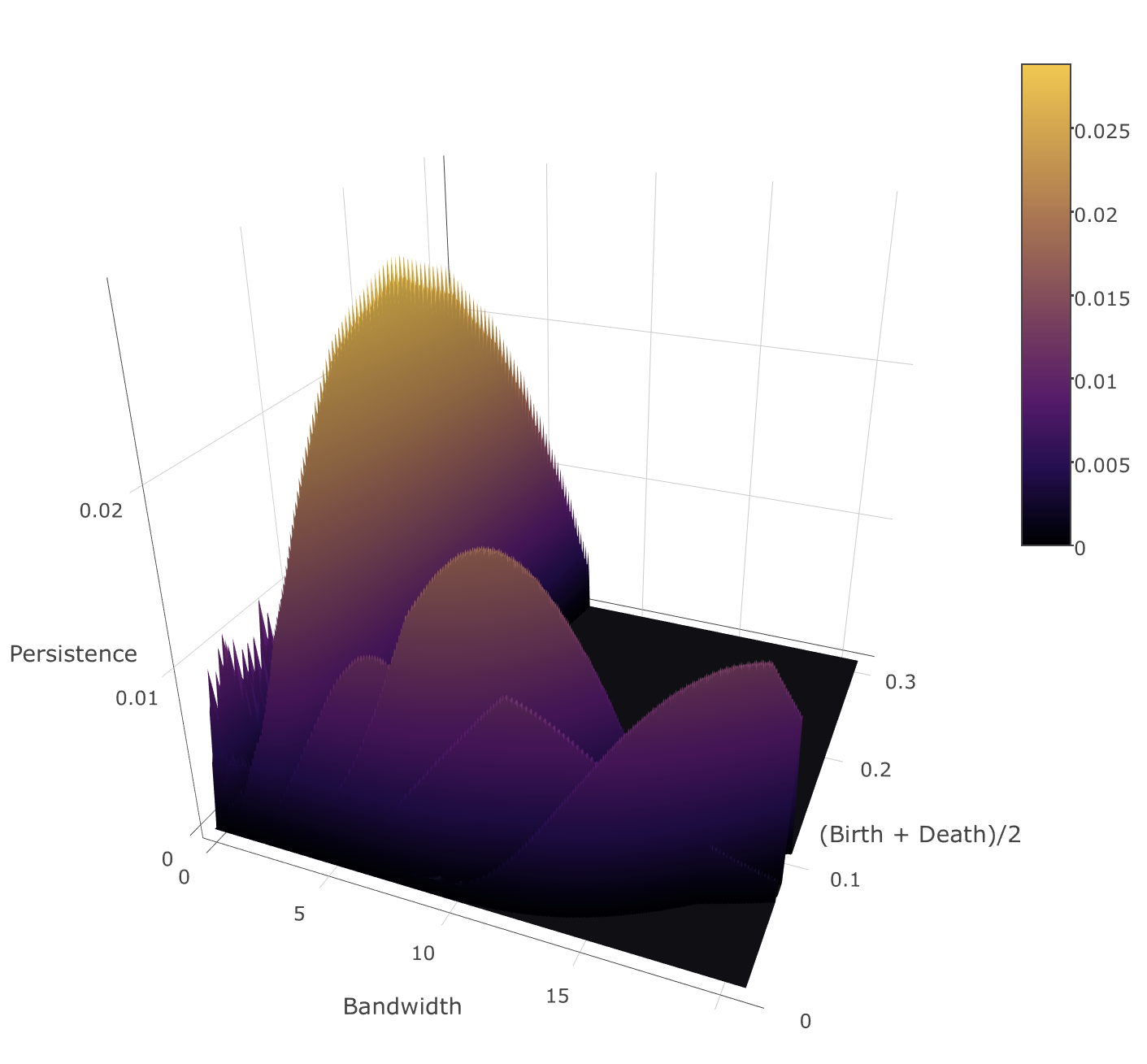}
\includegraphics[width=.45\linewidth]{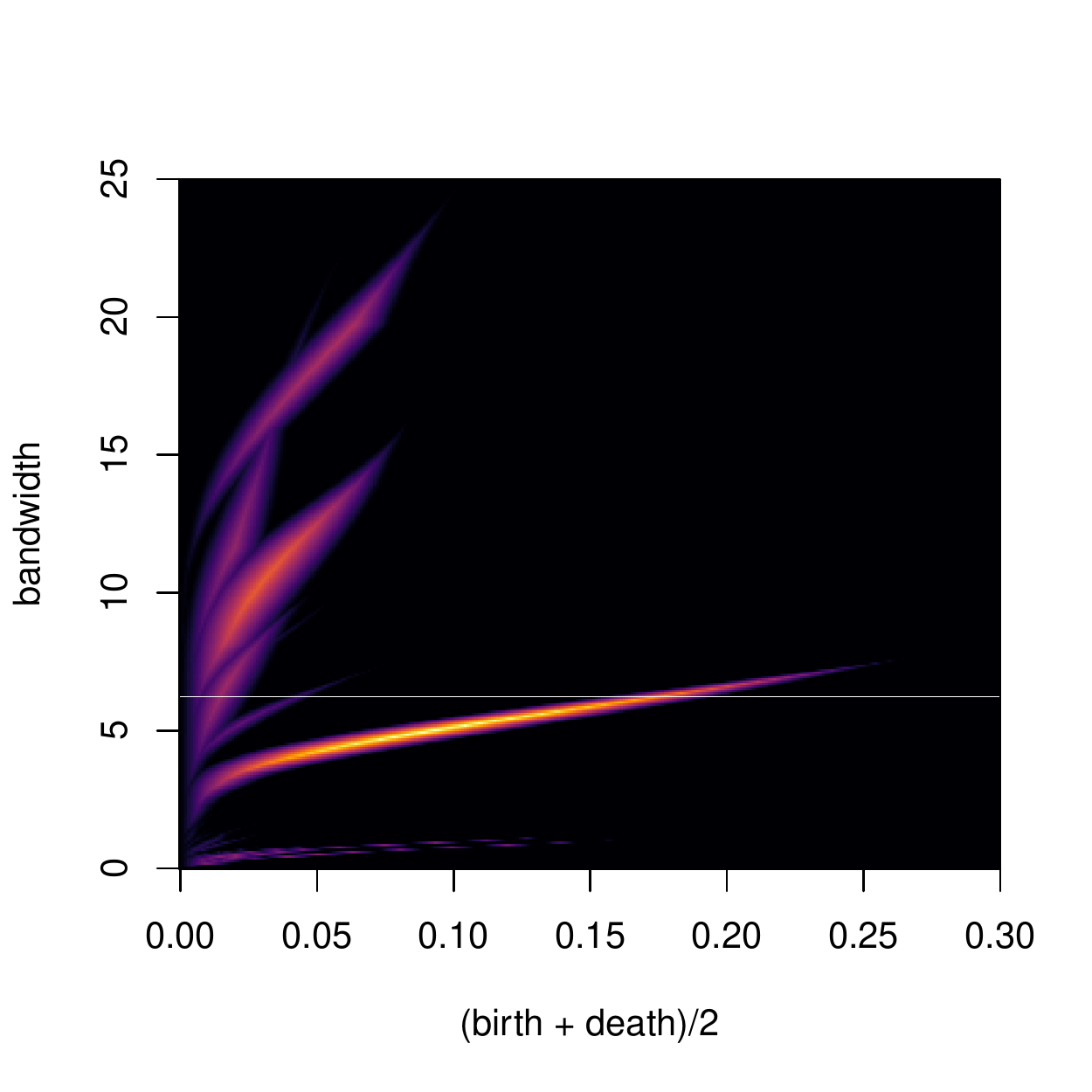}
\caption{Dimension $1$ Persistence Flamelets for earthquakes locations KDE (left) and its projection (right).}
\label{fig:quakesPLAIN}
\end{figure}
The dimension $1$ Persistence Flamelets of the kernel density estimator $\widehat{p}$ can be employed to assess whether or not kernel density estimators are able to recover these loops. The Persistence Flamelets shown in Figure \ref{fig:quakesPLAIN} presents $7$ crests, each of them representing one persistent loop in $\mathcal{F}$; this seems to suggest that at, different resolution, the kernel density estimator is able to recover all the $7$ main plates.  Notice that as opposed to the $0^{\tt th}$ dimensional case, where there is always one feature, the global maximum, dominating all the others, when analysing loops we can limit our analysis to the $1^{\tt st}$ Persistence Flamelets.

\begin{figure}
\centering
\includegraphics[width=.9\linewidth]{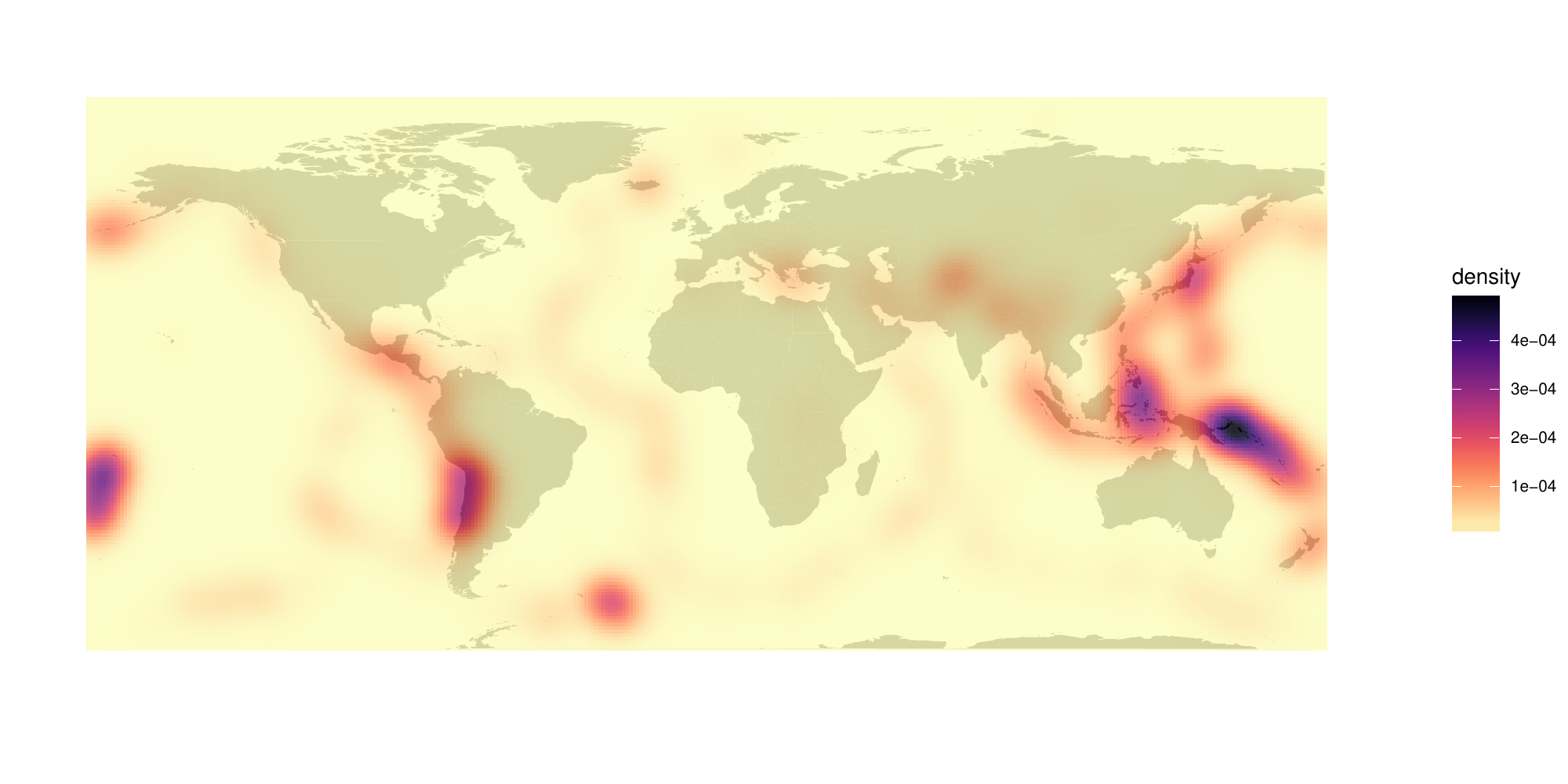}
\caption{Density estimation with the topologically aware bandwidth $\widehat{h}_{\text{TA}}$.}
\label{fig:quakesTB}

\includegraphics[width=.9\linewidth]{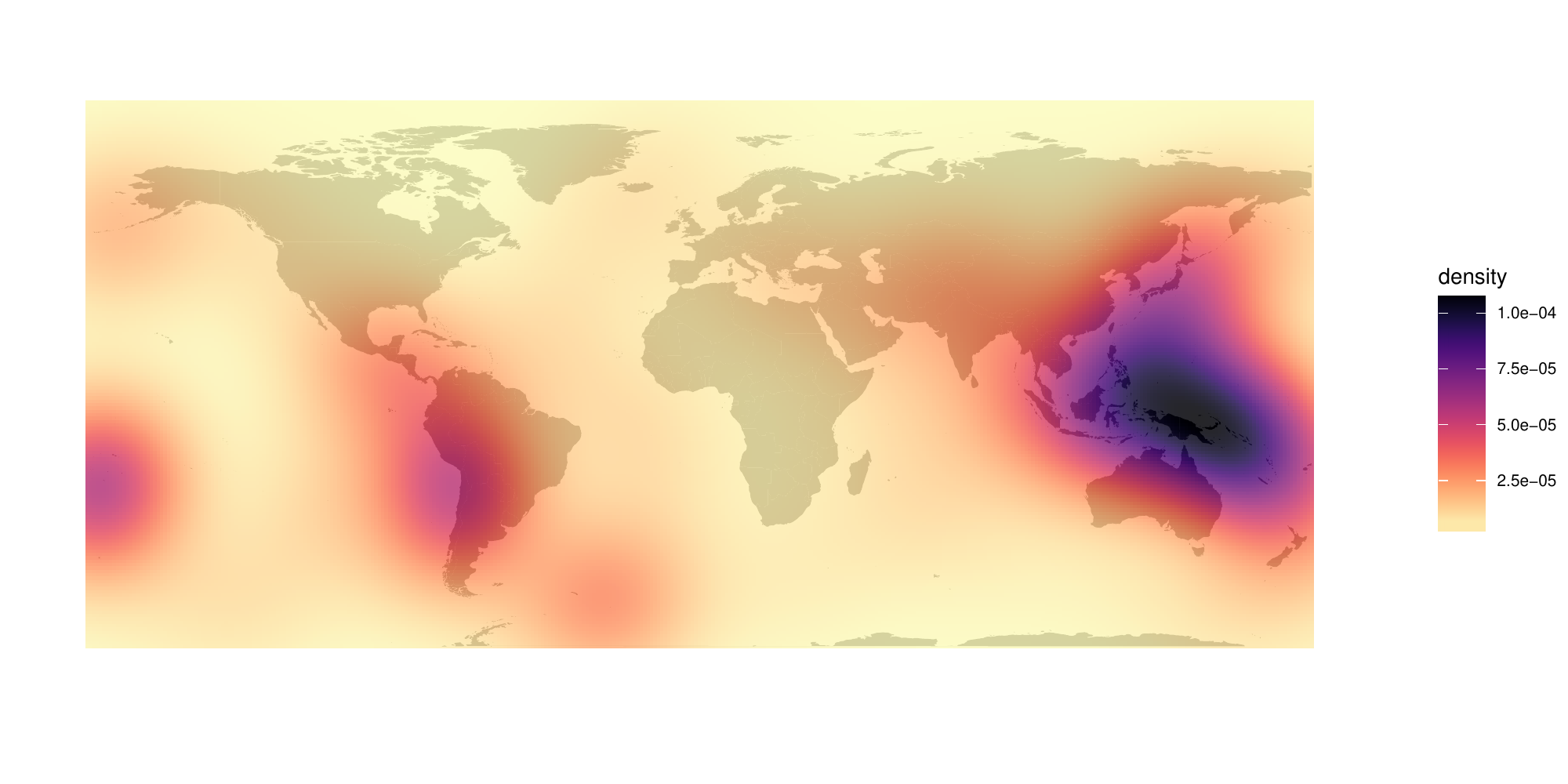}
\caption{Density estimation with extended Silverman Normal bandwidth $\widehat{h}_{\text{S}}$.}
\label{fig:quakesSB}

\includegraphics[width=.9\linewidth]{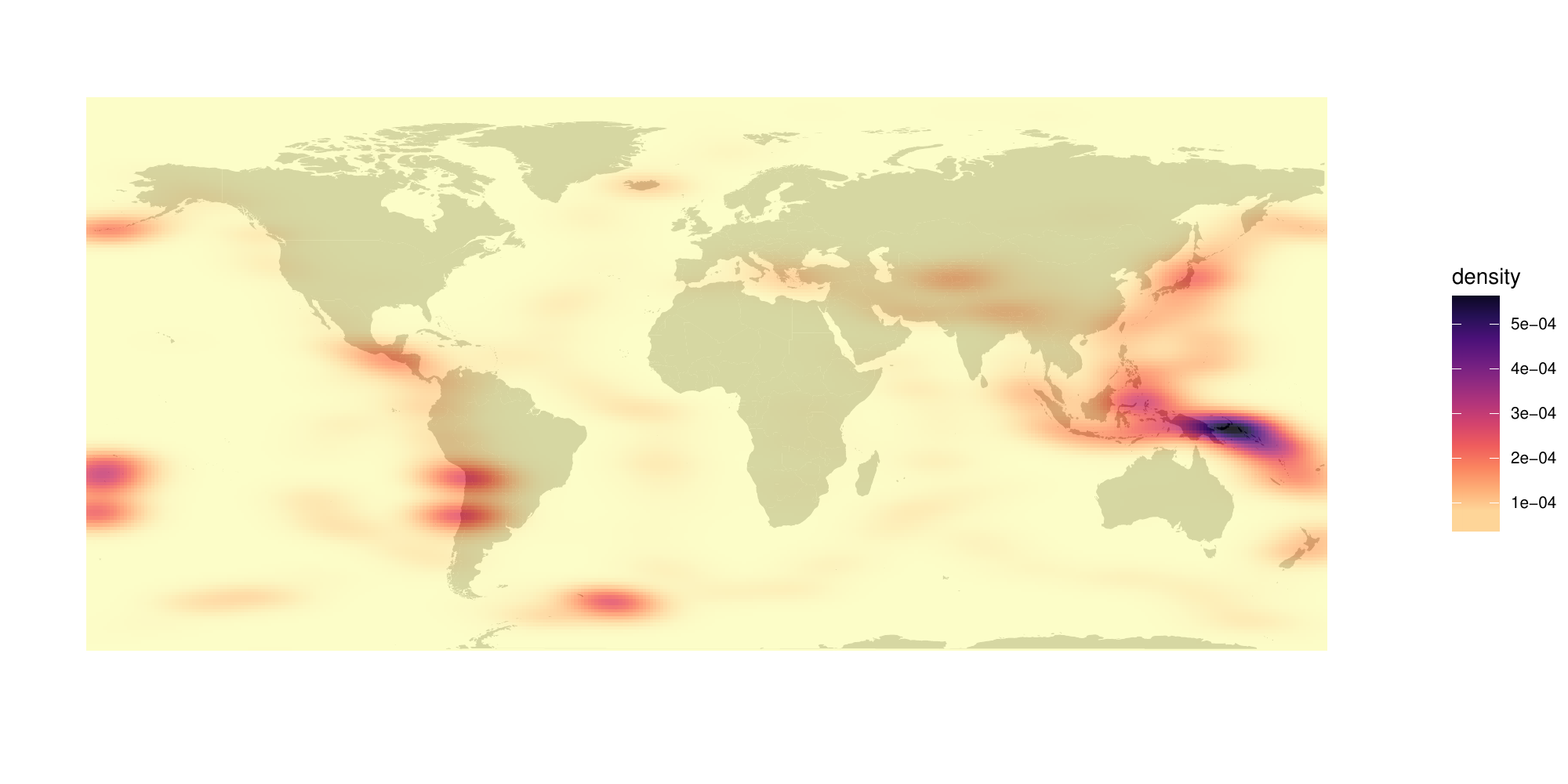}
\caption{Density estimation with anisotropic Plug--in bandwidth matrix $\widehat{\bm{H}}_{\text{PI}}$.}
\label{fig:quakesAB}
\end{figure}

If evaluating the importance of higher dimensional topological features such as loops is challenging from the point of view of exploration, this is even more true for the selection approach, where the topological structure is usually neglected (with the exception of local modes \cite{Genovese2016}). We argue that the Persistence Flamelets can be exploited in this task as well; intuitively, since persistence can be interpreted as a measure of the importance of each feature, bandwidths corresponding to peaks in the Persistence Flamelets result in estimators that highlight the most prominent features in the density. % Standard techniques for choosing the bandwidth such as cross validation, in fact, fail when the density is singular, i.e. concentrated around lower dimensional structures \ref{}, as in these cases. Since Persistence can be interpreted as a measure of the importance of each feature, we propose to use it as a heuristic to select the bandwidth. 

In this example specifically, the Persistence Flamelets shows that there is one loop that persists noticeably more than all the others; %As persistence is a measure of the importance of a feature, this means that there is one 
this suggests that there is one plate which is more neatly detected than all others. By selecting the value of $h$ that maximise the Persistence Flamelets, the \emph{topologically--aware} $\widehat{h}_{\text{TA}}$, we are forcing the density estimator to emphasize such feature. The kernel density estimator $\widehat{p}_{h_{\text{TA}}}$, shown in Figure \ref{fig:quakesTB}, is in fact concentrated on the contour of the Philippine plate, which is not surprising, since more than $26\%$ of the seismic activity in the given time interval was concentrated in the area between Philippine and Japan. 

To understand why such a topologically--aware bandwidth selection heuristic may be useful, let us compare it with more established methods for bandwidth selection: Silverman's Normal Rule and a Plug--in bandwidth selection criterion. We intentionally ignore cross validation methods because they have proven to fail when the density is singular, i.e. concentrated around lower dimensional structures \cite{Genovese2016}, as in these cases. 

The first alternative we consider is an extension of Silverman Normal Rule, one of the most famous ``rule of thumb'' for bandwidth selection, to the case of densities with singular features, as detailed in \cite{Genovese2017, Chacon2011}. More specifically, given a sample $\{\bm{X}_1, \ldots, \bm{X}_n\} \in \mathbb{R}^D$, from some distribution $P$, the optimal bandwidth $h$ for recovering the $d$--dimensional features is
\[
\widehat{h}_{\text{S}} = \left( \frac{4}{n(d+2)} \right)^{\frac{2}{4+d} }s,
\]
where $s=D^{-1} \sum_{j=i} ^D s^2_j$ and $s^2_j$ is the variance of the $j^{\tt th}$ variable. Despite the fact that we set $d = 1$, in order to take into account the loop structure, the density estimator, shown in Figure \ref{fig:quakesSB}, does not seem to recover any of the plates at all. 

The second approach we consider is a Plug--in bandwidth estimator $\widehat{\bm{H}}_{\text{PI}}$, obtained by minimizing the AMISE (Asymptotic Mean Integrated Square Error) w.r.t. the bandwidth $h$; details are given in \cite{Chacon2011}. Since limiting the case of scalar bandwidths, as we did until here, may seem too restrictive, in this final example we relax the hypothesis of spherical covariance and do not impose any structure on the bandwidth matrix $\bm{H}$. The additional complexity of the estimator does not however result in a better estimation: as we can see in Figure \ref{fig:quakesAB}, the plates structure of the true density is still not recognizable. 

\section{Discussion and Future Developments}

We have introduced a new multiscale topological summary, we have characterized it in a probabilistic framework and we have shown how to use to explore multidimensional time series and the relationship between the bandwidth and the topology of a kernel estimator. In the future we wish to exploit its good probabilistic properties to use it for statistical inference in addition to data description. More specifically, since we characterized the Persistence Flamelets in the context of multivariate time series, we plan to examine their use in testing for change point detection. 

Moreover we plan to investigate further the properties of Persistence Flamelets--related heuristics for bandwidth selection. We have already seen how picking the bandwidth that maximise the persistency seems to be promising, we plan to investigate it even further and to also consider using the Persistence Flamelets to select a bandwidth that reflects some previous knowledge on the topology of the object of interest.

Finally since we can think of the features that appears at many different resolution as the most relevant ones, we intend to explore persistence in bandwidth ranges as an additional measure of relevance for topological traits.

\medskip
\bibliographystyle{siam}
\bibliography{bibbi}

\end{document}